# RAPNet: A Receptive-Field Adaptive Convolutional Neural Network for Pansharpening

Tao Tang[a], Chengxu Yang[*]
*College of Computing, City University of Hong Kong, Hong Kong, China*
[a] tangtao.cs@my.cityu.edu.hk, [*]cyang326-c@my.cityu.edu.hk

*Abstract*—**Pansharpening, the fusion of a high-resolution panchromatic (PAN) image with a low-resolution multispectral (MS) image, is a crucial task in remote sensing. However, conventional Convolutional Neural Networks (CNNs) for this task are often limited by the translation-invariant nature of standard convolutions, which process all spatial locations with the same kernel, regardless of image content. To address this limitation, this paper proposes RAPNet, a novel content-adaptive convolutional network. The core of RAPNet is the Receptive-field Adaptive Pansharpening Convolution (RAPConv), a module that dynamically generates adaptive convolution kernels based on the local content of the input features. This allows the network to extract spatial details more effectively and adaptively. Furthermore, we introduce a Pansharpening Dynamic Feature Fusion (PAN-DFF) module, which utilizes an attention mechanism to adaptively balance the injection of spatial details and the preservation of spectral information. Extensive experiments on benchmark datasets demonstrate that the proposed RAPNet achieves excellent performance, outperforming existing methods both quantitatively and qualitatively. Ablation studies further validate the effectiveness of our proposed adaptive modules.**

## I. INTRODUCTION

Image fusion, particularly pansharpening in remote sensing, is a fundamental task in computer vision. It aims to generate high-resolution multispectral (HRMS) images by fusing panchromatic (PAN) and low-resolution multispectral (LRMS) data, which is critical for various downstream applications. Since the pioneering work of PNN, Convolutional Neural Networks have become the primary method for pansharpening due to their powerful ability to extract spatial features. Architectures like PanNet further advanced the field by introducing residual learning, enabling deeper and more effective models.

However, a key limitation of these methods is their reliance on standard convolutions, which are inherently translation-invariant. This means the same kernel is applied across the entire image, ignoring variations in local content and thus restricting the model's feature extraction capabilities. While effective, this content-agnostic approach is suboptimal for capturing the complex and diverse details in remote sensing imagery.

To address this challenge, we propose several innovations:

(1) A content-adaptive convolution method, RAPConv, that generates location-specific kernels to overcome the limitations of standard convolution, significantly enhancing spatial feature extraction.

(2) A dynamic feature fusion module, PAN-DFF, which uses an attention mechanism to adaptively balance spectral preservation and spatial detail injection.

(3) Comparative experiments to validate the effectiveness of the proposed network, RAPNet, offering a new and effective solution for the pansharpening task.

## II. RELATED WORK

In recent years, pansharpening based on deep learning has made significant progress. Researchers are no longer limited to traditional CNN architectures, but are exploring more diverse models. On the one hand, GANs are used to enhance the visual realism of fusion results, generating clearer texture details through adversarial training [1]. On the other hand, architectures represented by Transformers have attracted much attention due to their powerful global dependency modeling capabilities. For example, Li et al. [2] combined Transformers with deep unfolding networks to improve the performance and interpretability of the model. Ciotola et al. [3] proposed a novel unsupervised loss function that can jointly optimize the spectral and spatial fidelity of fused images without reference images. Zhou et al. [4] embedded a pre-trained masked autoencoder (MAE) as an image prior into the network, explicitly combining physical models with deep priors.

Despite advancements in network structures and learning paradigms achieved by the aforementioned methods, most of them still rely on a fundamental limitation: the "translation invariance" of standard convolution. Standard convolution shares the same kernel across all spatial positions in an image, and this content-agnostic approach overlooks the diversity and complexity of ground objects in remote sensing images, thereby limiting the flexibility of the model to adaptively extract and incorporate spatial details based on local content [5, 6].

To overcome this bottleneck, a cutting-edge research direction is to explore content-adaptive or dynamic convolution. These methods aim to dynamically adjust the parameters of convolutional kernels based on input features. Current explorations mainly fall into several categories: The first category is region-based adaptation, such as CANConv proposed by Duan et al. [6], which generates specific convolutional kernels for non-local regions with similar content in images, but makes adjustments at the region level, potentially ignoring pixel-level content changes within the same region. The second category involves changing the geometric shape of convolutional kernels. For example, ARConv proposed by Zhang et al. [7] can adaptively learn the rectangular size of convolutional kernels, but its adjustments mainly target the geometric shape of con-volutional kernels rather than the weights themselves. The third category utilizes dynamic convolution as an auxiliary module in large networks. For instance, Li et al. [5] integrated a dynamic high-pass filtering module into their SwinPAN model, but it serves a larger Transformer architecture.

These methods demonstrate the great potential of dynamically adjusting the convolution process. However, they either make adjustments at the regional level, focus on changing the shape of the convolution kernel, or use dynamic convolution as an auxiliary module. Currently, there is still a lack of a method that takes pixel-level content adaptation as the core mechanism and directly and flexibly generates corresponding kernel weights. The RAPConv proposed in this paper aims to fill this gap. The core idea of RAPConv is to dynamically generate unique kernel weights for each spatial location based on the feature information within its local receptive field. This enables our network, RAPNet, to extract and integrate spatial details in a more refined and adaptive manner, effectively overcoming the limitations of traditional convolution.

## III. METHOD

### A. *Overall Architecture*

The proposed RAPNet is an end-to-end convolutional neural network designed specifically for the pansharpening task. Its architecture is engineered to effectively extract and fuse spatial details from the high-resolution panchromatic (PAN) image and spectral information from the low-resolution multispectral (MS) image.

As illustrated in Figure 1, the network begins by processing the two inputs, PAN and MS, through a shared Edge Spatial Attention Module (ESAM) to adaptively enhance high-frequency edge information. Following this, the enhanced MS feature is up-sampled to match the spatial dimensions of the PAN feature. The two feature maps are then concatenated along the channel dimension and fed into the main backbone for deep spatial feature extraction. This backbone consists of an initial convolution, a series of stacked Receptive-field Adaptive Residual Blocks (RAP-ResBlocks), and a final convolutional layer. In the final stage, the rich spatial features learned by the backbone are fused with the up-sampled original MS image using a Pansharpening Dynamic Feature Fusion (PAN-DFF) module. This final module adaptively balances spatial and spectral information to reconstruct the high-quality, high-resolution multispectral (HRMS) output image.

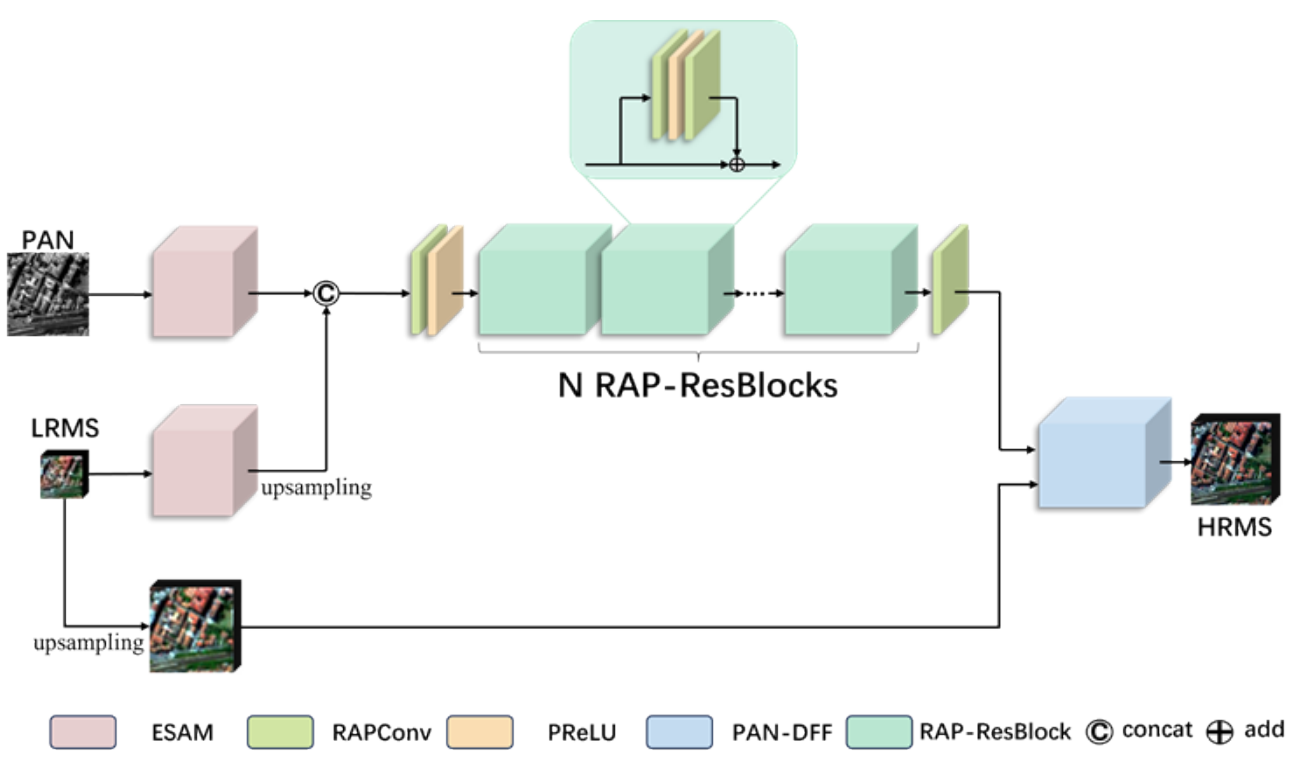

Figure 1. Overall structure of RAPNet

### B. *Receptive-Field Adaptive Convolution Kernel RAPConv*

This paper constructs an adaptive convolutional kernel RAPConv, which can adaptively adjust the convolutional kernel weights according to the local information of the input image to better extract spatial features in the image.

The reason why CNN structures can replace fully connected neural networks and achieve leapfrog progress in visual tasks is largely attributed to the "translation invariance" of the convolution operation. Thanks to this "translation invariance", CNNs greatly reduce the number of weight parameters, making training on large-scale visual datasets a reality and also reducing the risks of vanishing gradients and overfitting in the neural network.

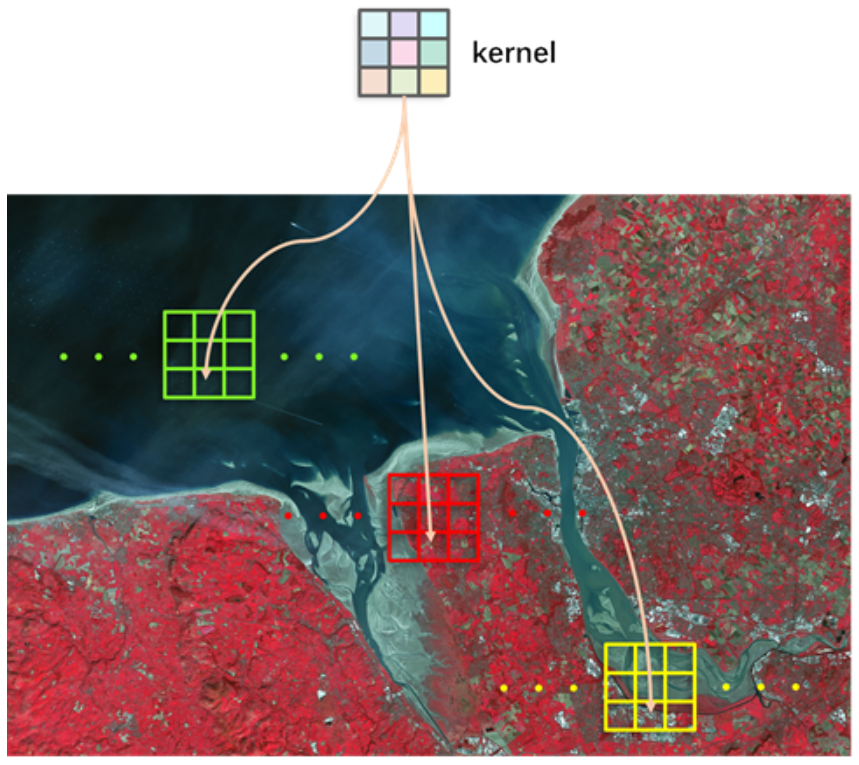

Figure 2. "Translation invariance" of traditional convolutional kernels

However, due to the inherent flaws in the "translation invariance" of convolutional computation, the further development of CNN models is some-what limited. As shown in Figure 2, during the smoothing process of the convolutional kernel on the image, the convolutional kernel parameters are the same at different positions.

In visual tasks, the content at different locations in an image contributes differently to the overall task. Traditional convolutional kernels only consider the position of pixels, without considering the image content itself (pixel value size), which limits the representational learning ability of neural networks. Therefore, it is necessary to construct a convolution kernel that is adaptive to the content at different positions in the image, as shown in Figure 3.

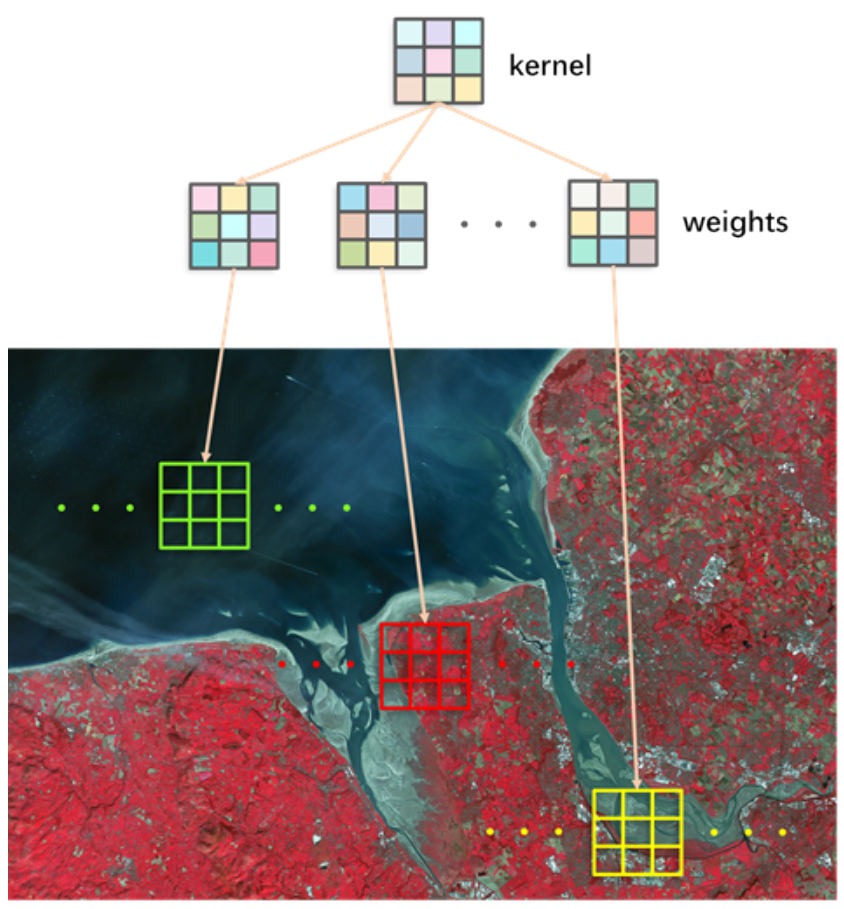

Figure 3. Adaptive convolutional kernel related to spatial context

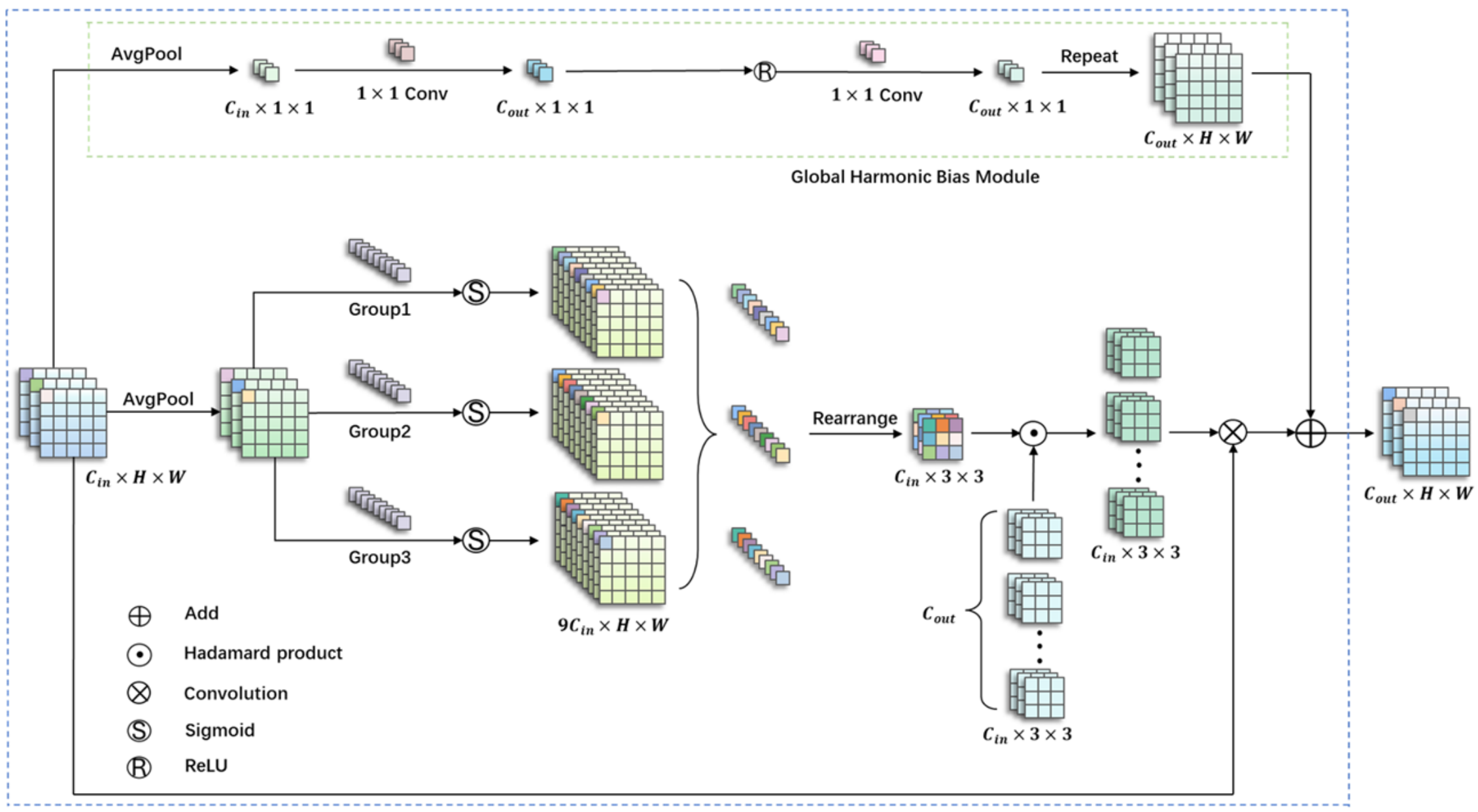

Figure 4. Schematic diagram of RAPConv structure

The features inputted into the kernel not only include the positional coordinates of pixels, but also the local spatial context information of the pixels themselves and even their receptive fields. Therefore, how to construct a convolution kernel that can effectively incorporate local spatial content information of images has become an urgent problem to be solved.

This paper proposes an adaptive convolution kernel, RAPConv for pansharpening inspired by Zhang et al [8]. By using a kernel with $1\times1$ size and a stride of 1 to perform convolution on each pixel, it extracts receptive field features at different spatial locations. Then, it performs Hadamard product with standard convolution kernels, effectively addressing the shortcomings brought by the "translation invariance" of traditional kernels and achieving adaptive learning spatial information. In addition, based on the characteristics of the pansharpening task, this paper also introduces the Global Harmonic Bias Module inspired by Jin et al. [9]. Its schematic diagram is shown in Figure 4 (taking $3\times3$ convolution as an example).

The structure of RAPConv can be primarily divided into two major components: receptive field spatial adaptive convolution and Global Harmonic Bias Module. First, let's introduce the receptive field spatial adaptive convolution, which can be further divided into two parts: receptive field local spatial feature extraction and adaptive convolution.

First, we introduce the method for extracting local spatial features from the receptive field. Assuming the size of the input image is $C\times H\times W$, where $C$ represents the number of channels in the input image, Figure 4 is drawn with $C=3$ as an example. We first introduce the method used for extracting local spatial information from the receptive field. For the original input image, a global average pooling is performed to generate a feature map with a size of $C\times H\times W$ (same size as the input image). Subsequently, a grouped convolution (there are a total of $9C$ $1\times1$ convolution kernels, which are evenly divided into $C$ groups, with 9 convolution kernels in each group. Essentially, each group of convolution kernels performs convolution on one channel of the input image) is applied to the feature map. The convolutional result is then input into the Sigmoid activation function, resulting in a feature map with a size of $9C\times H\times W$. For a convolution kernel with a size of $3\times3$, the receptive field size is $3\times3=9$ pixels. This design of grouped convolution effectively constructs a one-to-nine mapping, creating a convolution kernel that can adaptively learn the spatial context of local image regions while avoiding the heavy computational burden associated with manually extracting the receptive field spatial features of each pixel in the image (For example, the Unfold function in the Pytorch framework can be used to expand the feature image to manually extract receptive field spatial features, but it will undoubtedly incur heavy computational overhead).

The method of extracting local spatial features from receptive fields has been described above. Next, we will discuss how to incorporate the extracted spatial features from receptive fields into conventional convolution kernels, thereby generating kernels which can effectively extract local spatial features.

As mentioned earlier, for each pixel point in the input image, nine-pixel points are generated to represent its receptive field spatial features. To inject these features into a convolution kernel of size $3\times3$, the rearrange function from the einops library is needed to change its shape to the corresponding $3\times3$ size. For an input image with $C$ channels, this is equivalent to generating an attention weight of size $C\times3\times3$. By performing Hadamard product with a regular convolution kernel of shape $C\times3\times3$, the receptive field spatial attention weights can be injected into the regular convolution kernel. This means that when the convolution kernel slides over the input image, the weights at different spatial positions are different. If an output image with $P$ channels is ultimately required, $P$ regular convolution kernels need to be set accordingly. It should be noted that Figure 4 is drawn using $P=3$ as an example.

Since dynamic convolution techniques, represented by adaptive convolution, aim to improve the performance of neural networks by focusing on spatial local features in feature images, they may lead to spatial distortions in pansharpening tasks due to neglecting global information. In the RAPConv constructed in this paper, the Global Harmonic Bias Module (GHBM) is introduced to enhance the convolution kernel's ability to extract global information. As shown in Figure 4, the GHBM first performs global average pooling on the input feature map of the convolution kernel, outputting a feature map of size $C\times1\times1$. Then, it performs another $1\times1$ convolution, activates it using the ReLU activation function, and finally performs another $1\times1$ convolution to output a feature map of size $P\times1\times1$. Finally, the Repeat function in Pytorch is used to generate a feature map of size $P\times H\times W$, which is added to the feature map of size $P\times H\times W$ generated by the receptive field space adaptive convolution part introduced earlier as the final output result of the RAPConv convolution kernel, with its size consistent with the input of the RAPConv convolution kernel.

### C. Residual Block based on RAPConv

Based on the adaptive convolution RAPConv described above, this paper constructs a residual module, RAP-ResBlock.

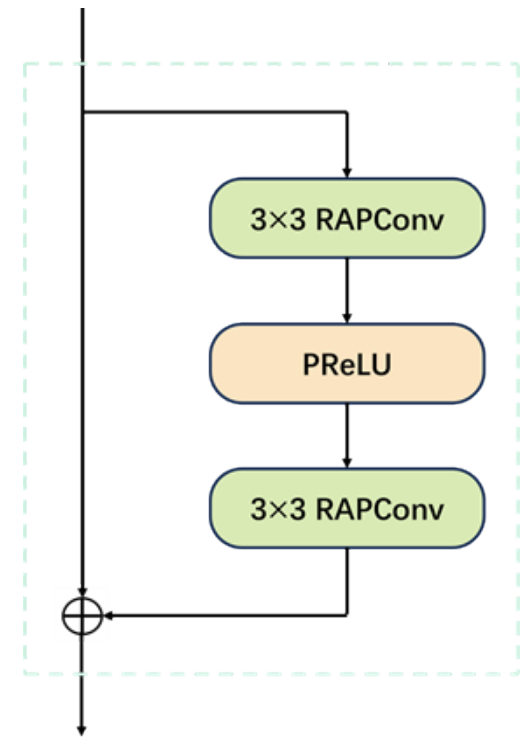

Figure 5. The structure of RAP-ResBlock

As shown in Figure 5, the input of the RAP-ResBlock module first passes through an adaptive convolution kernel of size $3\times3$ in RAPConv, is activated by a PReLU activation function, and finally obtains the final result by passing through another adaptive convolution kernel of size $3\times3$ in RAPConv and adding the original input transmitted through a skip connection. In RAPNet, there are a total of four RAP-ResBlocks.

### D. A Spatial-Spectral Dynamic Feature Fusion Mechanism

Traditional networks for pansharpening tasks mostly adopt a unified structure, which upsamples the MS from satellites to the same size as the PAN, and then directly adds them to the output from the spatial detail learning network, resulting in the final result. However, this method may not be able to achieve a good balance between spectral fidelity and spatial detail injection.

Inspired by Yang et al [10], we design a dynamic feature fusion mechanism based on attention and $1\times1$ convolutions, named PAN-DFF, which is suitable for pansharpening tasks. It can adaptively fuse the output of the spatial detail learning network with the low spatial resolution multispectral images (LRMS) obtained by upsampling the original multispectral images (MS) from satellites to the size of the PAN image, thus achieving a good balance between spatial detail injection and spectral fidelity, and adaptively highlighting the most important spatial and spectral features according to a dynamic selection mechanism. The structure of PAN-DFF is shown in Figure 6.

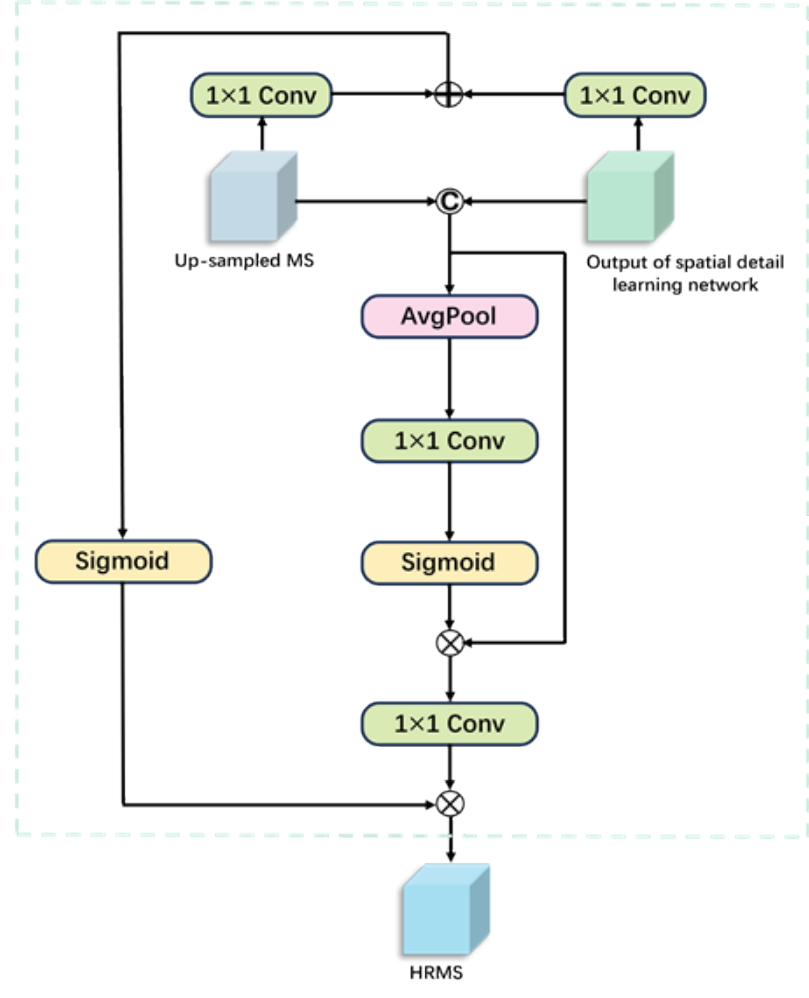

Figure 6. The structure of PAN-DFF

## IV. EXPERIMENTS

### A. Experimental Setup

We conducted experiments using PyTorch 1.10.0 and NVIDIA GeForce RTX 4090 (24G). We trained for 500 epochs with a batch size of 32, using Adam as the optimizer and a learning rate of 0.00025. The loss function employed was MSE.

### B. Dataset, Metrics and Baseline

We use the publicly available dataset PanCollection, which was constructed by Deng et al. [11], and includes satellite images from WorldView-3, QuickBird, GaoFen2. Deng et al. [11] detailed 3 non-deep learning methods: BDSD-PC, MTF-GLP-FS, and BT-H, as well as 3 deep learning methods: PNN, PanNet, and FusionNet. We conducted a fair comparison with RAPNet on the unified public dataset PanCollection and compared them using image fusion quality evaluation metrics such as ERGAS, SAM, Q8, and SCC [11]. The results are as follows.

### C. Test Results of WorldView-3 Simulated Dataset

This method was tested alongside other methods on the WorldView-3 simulation test dataset, consisting of 20 test images, each with a size of $256\times256\times8$. The test results are presented in Table 1 (bold font indicates the best data, and an underscore indicates the second-best data, the same below).

Table 1 Test Results of the WorldView-3 Simulation Dataset

| Method/Indicator | ERGAS | SAM | Q8 | SCC |
|---|---|---|---|---|
| BDSD-PC | 4.698±1.617 | 5.429±1.823 | 0.829±0.097 | 0.908±0.040 |
| MTF-GLP-FS | 4.701±1.597 | 5.316±1.766 | 0.833±0.092 | 0.901±0.045 |
| BT-H | 4.579±1.495 | 4.920±1.425 | 0.832±0.094 | 0.925±0.024 |
| PNN | 2.696±0.675 | 3.917±0.789 | 0.887±0.095 | 0.973±0.009 |
| PanNet | 2.675±0.686 | 3.845±0.713 | 0.889±0.092 | 0.974±0.009 |
| FusionNet | 2.492±0.633 | 3.372±0.706 | 0.899±0.089 | 0.979±0.007 |
| RAPNet | **2.353±0.629** | **3.369±0.723** | **0.902±0.091** | **0.982±0.008** |
| Optimal Value | 0 | 0 | 1 | 1 |

From the various evaluation indicators and their visualizations in the test results, it can be seen that RAPNet achieved the best fusion effect. The results of applying these methods to a multispectral image in the test set are visualized in Figure 7.

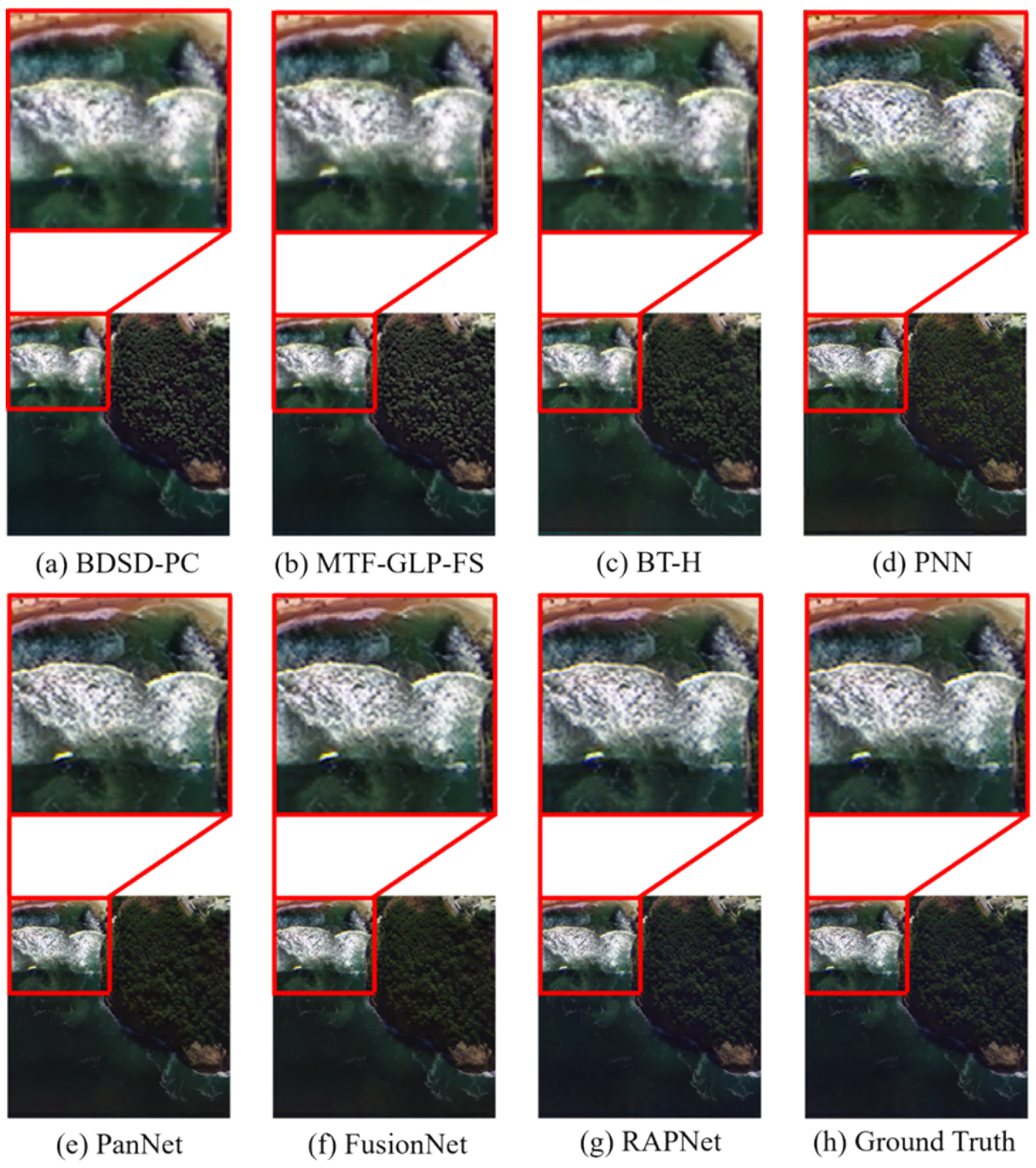

(a) BDSD-PC (b) MTF-GLP-FS (c) BT-H (d) PNN

(e) PanNet (f) FusionNet (g) RAPNet (h) Ground Truth

Figure 7. Visualization of the test results of the WorldView-3 simulation dataset

### D. Test Results of WorldView-3 Real Dataset

To further verify the fusion performance of the method proposed in this paper, it was compared with other pansharpening methods on the WorldView-3 real test dataset, and the results are shown in Table 2.

Table 2. Test Results Of The WorldView-3 Real Dataset

| Method/Indicator | $D_\lambda$ | $D_S$ | $QNR$ |
|---|---|---|---|
| BDSD-PC | 0.0625 ± 0.0235 | 0.0730 ± 0.0356 | 0.8698 ± 0.0531 |
| MTF-GLP-FS | 0.0197 ± 0.0078 | 0.0630 ± 0.0289 | 0.9187 ± 0.0347 |
| BT-H | 0.0425 ± 0.0139 | 0.0754 ± 0.0328 | 0.8857 ± 0.0431 |
| PNN | 0.0232 ± 0.0095 | 0.0461 ± 0.0159 | 0.9319 ± 0.0204 |
| PanNet | **0.0183 ± 0.0059** | 0.0477 ± 0.0203 | 0.9349 ± 0.0206 |
| FusionNet | 0.0246 ± 0.0087 | 0.0392 ± 0.0153 | 0.9363 ± 0.0198 |
| RAPNet | 0.0191 ± 0.0063 | **0.0389 ± 0.0163** | **0.9374 ± 0.0201** |
| Optimal Value | 0 | 0 | 1 |

The results reflect the excellent performance of RAPNet in spatial detail learning tasks, while there is still room for improvement in spectral information extraction.

### E. Ablation Experiment

This paper compares two different network structures under the same WorldView-3 simulation dataset. One of the network structures is the RAPNet proposed in this paper, while the other is a structure obtained by replacing all RAPConv in RAPNet with normal convolutional kernels. The test results of the two structures under the same conditions are shown in Table 3.

Table 3. Comparison With and Without RAPConv

| Method/Indicator | ERGAS | SAM | Q8 | SCC |
|---|---|---|---|---|
| No RAPConv | 2.639±0.630 | 3.764±0.711 | 0.891±0.091 | 0.974±0.007 |
| RAPConv | **2.353±0.629** | **3.369±0.723** | **0.902±0.091** | **0.982±0.008** |
| Optimal Value | 0 | 0 | 1 | 1 |

As can be seen from the table above, after adding the adaptive convolution RAPConv, the network has achieved improvements in various evaluation metrics, indicating that the adaptive convolution RAPConv can effectively enhance the performance of the network in pansharpening tasks.

## V. CONCLUSION

In this paper, we introduced RAPNet, a novel network architecture designed for pansharpening. Our approach tackles the inherent "translation-invariance" limitation of standard convolutions by proposing the RAPConv. This module dynamically generates context-aware kernels to enhance spatial feature extraction. Furthermore, our PAN-DFF module adaptively balances spectral fidelity and spatial detail injection. The results validate that our method can effectively fuses images and achieves superior performance compared to other approaches. For future work, we will explore several advanced directions which include leveraging generative diffusion models for unsupervised learning, developing physics-informed networks constrained by sensor properties to improve robustness.